\documentclass[acmtog]{acmart}
\renewcommand\footnotetextcopyrightpermission[1]{} % removes footnote with conference information in first column
\usepackage{pifont}
\usepackage{graphicx}
\usepackage{subcaption}
\usepackage{tablefootnote}
\AtBeginDocument{%
  \providecommand\BibTeX{{%
    \normalfont B\kern-0.5em{\scshape i\kern-0.25em b}\kern-0.8em\TeX}}}

\setcopyright{acmcopyright}
\copyrightyear{2019}
\begin{document}

\title{IIITM Face: A Database for Facial Attribute Detection in Constrained and Simulated Unconstrained Environments}

\author{Raj Kuwar Gupta}
\authornotemark[1]
\authornotemark[2]
\email{rajkuwargupta1996@gmail.com}

\author{Shresth Verma}
\email{vermashresth@gmail.com}
\orcid{0000-0003-0370-5471}
\authornote{All these authors are from the same institution.}
\authornote{Equal Contribution}
\author{KV Arya}
% \affiliation{\institution{ABV-Indian Institute of Information Technology and Management Gwalior, India-474015}}
\authornotemark[1]

\orcid{0000-0001-7117-1745}
\email{kvraya@iiitm.ac.in}

% \affiliation{\institution{ABV-Indian Institute of Information Technology and Management Gwalior, India-474015}}
% \affiliation{\institution{Indian Institute of Information Technology and Management Gwalior, India-474015}}

\author{Soumya Agarwal}
\authornotemark[1]
\affiliation{\institution{ABV-Indian Institute of Information Technology and Management Gwalior, India-474015}}
\email{soumya.agarwal@gmail.com}

\author{Prince Gupta}
\affiliation{\institution{Netaji Subhas University of Technology, Delhi, India-110078}}
\email{guptaprince223@gmail.com}

\renewcommand{\shortauthors}{Gupta and Verma, et al.}

\begin{abstract}
This paper addresses the challenges of face attribute detection specifically in the Indian context. While there are numerous face datasets in unconstrained environments, none of them captures emotions in different face orientations. Moreover, there is an under-representation of people of Indian ethnicity in these datasets since they have been scraped from popular search engines. As a result, the performance of state-of-the-art techniques can't be evaluated on Indian faces. In this work, we introduce a new dataset \textbf{IIITM Face} for the scientific community to address these challenges. Our dataset includes 107 participants who exhibit 6 emotions in 3 different face orientations. Each of these images is further labelled on attributes like gender, presence of moustache, beard or eyeglasses, clothes worn by the subjects and the density of their hair. Moreover, the images are captured in high resolution with specific background colors which can be easily replaced by cluttered backgrounds to simulate `in the Wild' behavior. We demonstrate the same by constructing IIITM Face-SUE. Both IIITM Face and IIITM Face-SUE have been benchmarked across key multi-label metrics for the research community to compare their results. 
\end{abstract}





\maketitle

\section{Introduction}
Face-attribute detection, which is aimed at identifying all the facial attributes from a given image, is a classical problem in the domain of multi-label classification. These attributes provide essential information about mid-level representations of faces that are abstracted between very low pixel level features and high level identity labels. The attributes can be very diverse and can include, for instance, gender, presence or absence of facial hair like beard or moustache, color and density of hair etc. Reliable identification of these attributes is crucial so as to have an intuitive and human interpretable face description. Moreover, accurately recognizing these attributes can play an important role in designing Human Computer Interaction (HCI) systems which need to be aware of gender and emotion of the user to respond appropriately.

% \begin{figure}[htp]
% \centering
%   \begin{subfigure}[b]{width=6cm}
%   \includegraphics[width=\textwidth]{download_marked.png}
% %   \caption{Steps for simulating unconstrained environment}
%     \caption{}
%   \label{fig:Ng1} 
% \end{subfigure}
% \begin{subfigure}[b]{width=6cm}
%   \includegraphics[width=\textwidth]{download_em_pose_final.png}
%   \caption{}
%   \label{fig:Ng2}
% \end{subfigure}
% \caption{(a) Steps involving the generation of simulated unconstrained environment database. (b) 'In the wild' images for a single subject across all poses and emotions}
% \end{figure}
% \vspace{-1mm}
\begin{figure}[h]
    \centering
    \includegraphics[width=6.5cm]{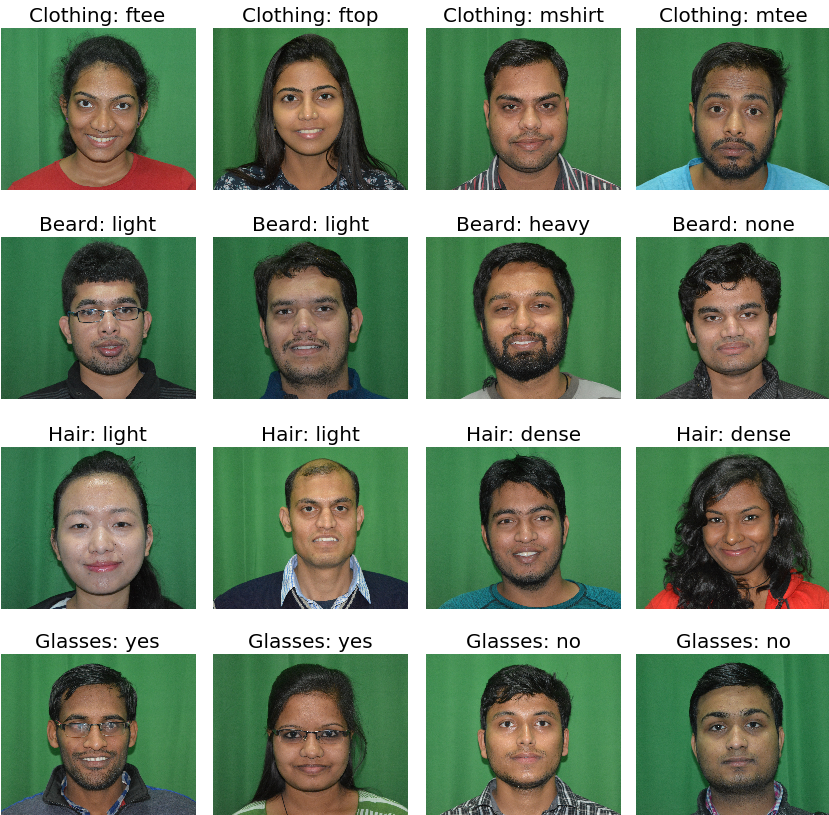}
    \vspace{-6pt}
    \caption{Attribute diversity in IIITM Face Dataset}
    \label{fig:constant}
\end{figure}
% \vspace{-5mm}
\begin{figure}[h]
    \centering
    \includegraphics[width=6.5cm]{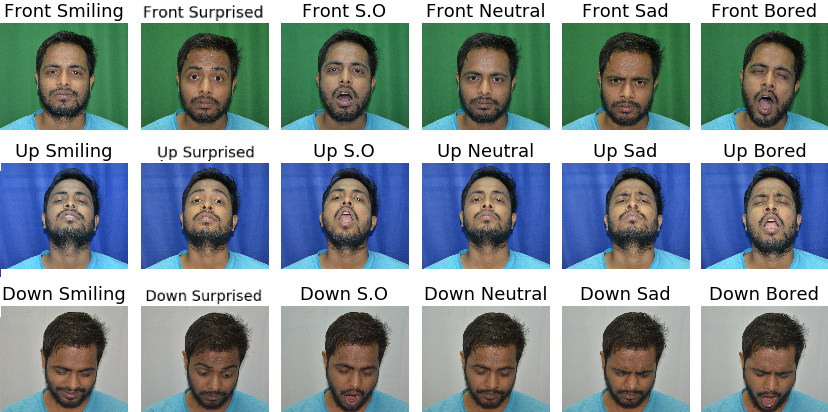}
    \vspace{-6pt}
    \caption{Pose and Emotion variation in IIITM Face Dataset }
    \label{fig:posevar}
\end{figure}
% \vspace{-2mm}
While there are several datasets  available for face-attribute detection on faces in the Wild, they lack simultaneous variability in poses and emotions as they are obtained by scraping from popular search engines \cite{celebA,lfwdataset}. There is also a question of variability of backgrounds in these images as they are often close-up shots of faces. Additionally, the research community also feels an insufficiency of data of Indian faces. 

%Mention about the IITK dataset for face recognition and the indian movie star dataset.

To address these challenges we have collected a dataset - IIITM Face Dataset which is being released with this work. IIITM Face has been constructed by the participation of 107 students and staff at ABV-IIITM Gwalior. 
%Each participant has exhibited 6 emotions in 3 different orientations. The total number of images are 1928. 
The images in this dataset have captured subjects in all possible combinations of 6 emotions and 3 different orientations (Fig. \ref{fig:posevar}). Along with this, constant facial attributes of the subjects are also marked (Fig. \ref{fig:constant}). The detailed description of the dataset across all the attributes has been provided in Table \ref{stats-of-database}.
% been constructed by the participation of 107 students and staff atABV-IIITM Gwalior. All these images have been labelled to captureface-attributes like presence of mustache or eyeglasses, presenceand density of beard and hair, gender of the subject along with whathe/she is wearing. The detailed description of the dataset across allthe 8 attributes has been provided in Tab

The classification of these facial attributes has been achieved by using three classifiers - Logistic Regression (LR), Support Vector Machine (SVM) \cite{svm} and ResNet \cite{resnet}. The implementation details and hyperparameters used are outlined in section \ref{experimental-setup}.
All the images in this dataset are captured in high resolution with specific background colors which can easily be replaced by cluttered backgrounds to simulate `in the Wild' behavior. We use complex scenes from \cite{shi2015hierarchical} and use them as backgrounds in order to construct a new dataset, IIITM Face Simulated Unconstrained Environment (hereafter referred to as \textbf{IIITM Face-SUE}, see Fig. \ref{fig:aug}). All the faces in IIITM Face-SUE are subjected to translation to ensure that learning is position invariant. Our experiments on IIITM Face-SUE show that there is a significant performance degradation across all metrics, thereby presenting an opportunity for exploring approaches which are invariant to the background noise.
%and thereby presents an opportunity for the research community to develop approaches which are invariant to the background noise.
%All the 3 classification models are then again trained only on this distorted dataset and their performance is evaluated. The results are compared across key metrics drawn from.
% refer the ICML paper here and refer the results section.
 
The contributions of this work are summarized as follows - (i) We release an extensively labelled face-attribute dataset, i.e., IIITM Face Dataset with Indian Faces. (ii) We perform rigorous benchmarking of this dataset across key metrics of multi-attribute classification. (iii) We compare the classification performance on the IIITM Face and IIITM Face-SUE datasets.
\vspace{-3mm}

\section{Related Work}
In the early days of research in face-attribute detection,  the dataset sizes were small (about 500-1000 images from 20-120 participants) and had relatively fewer attributes as is evident in \cite{YaleFace, umistDatabase, humanscandatabase, DatabaseofFaces} \nocite{PhillipsWHR98}. 
%These datasets were quite small in size (about 500-1000 images from 20-120 participants) and had relatively few attributes as the primary objective was the identification of faces. But these were aptly suited for the computational power available then. However, as algorithms improved after receiving boost from the improvement in computation power available, these algorithms grew data hungry. 
Recently, larger datasets have been constructed given the widespread availability of data. One of the first such datasets was Labelled Faces in the Wild (LFW) dataset \cite{lfwdataset} which contained 13233 images scraped from the internet identifying 5749 people. In recent research, CelebA \cite{celebA} dataset has been used dominantly due to its sheer size of 200K celebrity images, each with 40 attribute annotations.
%LFW+ dataset ??.
Another dataset, VGGFace \cite{parkhi2015deep}, containing 2.6M images across 2,622 people is one of the largest publicly available dataset. The size of this dataset was further increased and released as VGGFace2 Dataset \cite{cao2018vggface2} containing 3.3M images of 9,131 people. VGGFace2, along with UMDFaces \cite{bansal2017umdfaces} are among very few publicly available large scale datasets containing pose information corresponding to each face image instance.

%Another class of datasets focus entirely on facial emotion detection. 

However, in the Indian context, there has been a relative dearth of data. Racial diversity is an element that is crucial for making facial recognition systems more reliable. For instance, VGGFace2 \cite{cao2018vggface2} and FairFace\cite{FairFace} focused on increasing the proportion of ethnically diverse faces and faces from different professions. The earliest dataset on Indian faces is the Indian Face Dataset \cite{iitk2002ho}. It contains 11 images of 40 different people containing pose and emotion attributes. There are also other datasets such as Indian Face Age Database (IFAD) \cite{sharma2015indian} which contains attribute information of Indian celebrities but has limited number of subjects and low resolution images. Another dataset, Indian Spontaneous expression \cite{happy2015indian}, presented video frames of subjects while watching emotional video clips thus giving rich information on emotion transitions. One large scale dataset available for Indian personalities is Indian Movie Face Database \cite{indianceleb} which contains 34512 images of 100 Indian celebrities with attribute labels consisting of age, pose, gender, expression and amount of occlusion. But all these datasets on Indian faces are limited in either the resolution of images, diversity in attributes or size of the dataset.

One major class of problems that is often studied on these datasets is facial attribute classification\cite{liu2015deep, taherkhani2018deep, sun2018deep}. This can be posed as a multi task joint learning problem or can be learnt through an ensemble of models for every attribute. Joint learning has the advantage that it can discover relationships across attributes and this is especially useful since facial attributes are highly correlated. On the other hand, separate models fail when large facial variations are present. In our work, we will be exploring both ensemble of models approach and multi task joint learning on our presented dataset.

\section{Methodology}
\subsection{Overview of IIITM Face Dataset}
The IIITM Face Dataset contains 1928 images of 107 people spanning across a wide range of facial attributes such as pose, gender, emotion, facial hair (beard and moustache), glasses, hair and clothing. The subjects of the dataset are students and staff members of IIITM institute and there are at least 18 images for each subject.  While the facial attributes and apparel labels are constant across one subject, the pose and emotion attributes cover all possible combinations.

%The dataset is designed to serve as a benchmark for facial matching tasks across poses and emotions and also for facial attribute classification tasks. More specifically, it is designed keeping in mind the lack of high resolution facial data from Indian context. Most large scale datasets have either very low number of subjects of Indian ethnicity or the images are scraped from internet and thus are not in high resolution. 

\begin{table}[h]
\footnotesize
\centering
\begin{tabular}{|c|c|c|c|c|c|c|}
\hline
Dataset & \# Subjects & \# Images & Resolution & Emotion & Pose & \begin{tabular}[c]{@{}c@{}}Facial\\ Attributes\end{tabular} \\ \hline
IITK Face & 40 & 440 & 640X480\tablefootnote{All images are in grayscale} & Yes & Yes & No \\ \hline
IMFDB & 100 & 34512 & Varying\tablefootnote{Small images in the range of 100-350 pixels in width and height} & Yes & Yes & Yes \\ \hline
IFAD & 55 & 3296 & 128X128 & Yes & Yes & No \\ \hline
ISED & 50 & 428\tablefootnote{Videos of the participants at 50 frames per second} & 1920X1080 & Yes & No & No \\ \hline
IIITM Face & 107 & 1928 & 2992X2000 & Yes & Yes & Yes \\ \hline
\end{tabular}
\caption{Comparison of Indian Face Databases}
\label{tab:comparison}
\vspace{-6mm}
\end{table}

Table \ref{tab:comparison} compares IIITM dataset with other publicly available facial image datasets in Indian context. Note that this dataset isn't meant to serve as a `in the Wild' (unconstrained environment) dataset. However, in section \ref{IIITM-Face-SUE} we describe how we can post-process this dataset to treat it as a `in the Wild' dataset.

\subsection{Attribute Descriptions}
For each face we, manually label it for the following eight attributes:
\begin{itemize}
    \item Pose: Front, Up and Down
    \item Gender: Female, Male
    \item Emotion: Neutral, Sad, Smiling, Surprised, Surprised with Open Mouth (S.O), Bored
    \item Beard: Heavy, Light, None
    \item Moustache: Yes, No
    \item Glasses: Yes, no
    \item Hair: Light, Dense
    \item Cloth: Female T-Shirt(F. Tee), Female Top (F. Top), Male Shirt (M. Shirt), Male T-Shirt (M. Tee) 
\end{itemize}
% \vspace{-2mm}
\begin{table}[h]
\footnotesize
\begin{tabular}{ccccc}
\hline
\multicolumn{5}{|c|}{\textbf{Attributes}}                                                                                                                                                            \\ \hline
\multicolumn{1}{|c|}{\textbf{Orientation}} & \multicolumn{1}{c|}{\textbf{Gender}}  & \multicolumn{2}{c|}{\textbf{Emotion}}                                     & \multicolumn{1}{c|}{\textbf{Beard}} \\ \hline
\multicolumn{1}{|c|}{Down: 642}            & \multicolumn{1}{c|}{Female : 360}     & \multicolumn{1}{c|}{Neutral : 322} & \multicolumn{1}{c|}{Surprised : 322} & \multicolumn{1}{c|}{Heavy : 288}    \\ \hline
\multicolumn{1}{|c|}{Up: 642}              & \multicolumn{1}{c|}{Male : 1568}      & \multicolumn{1}{c|}{Sad : 321}     & \multicolumn{1}{c|}{S.O : 321}       & \multicolumn{1}{c|}{None : 792}     \\ \hline
\multicolumn{1}{|c|}{Front : 644}          & \multicolumn{1}{c|}{}                 & \multicolumn{1}{c|}{Smiling : 321} & \multicolumn{1}{c|}{Bored : 321}     & \multicolumn{1}{c|}{Light : 848}    \\ \hline
\multicolumn{5}{c}{}                                                                                                                                                                                 \\ \hline
\multicolumn{1}{|c|}{\textbf{Moustache}}   & \multicolumn{1}{c|}{\textbf{Glasses}} & \multicolumn{2}{c|}{\textbf{Cloth}}                                       & \multicolumn{1}{c|}{\textbf{Hair}}  \\ \hline
\multicolumn{1}{|c|}{Yes : 1135}           & \multicolumn{1}{c|}{Yes : 180}        & \multicolumn{1}{c|}{F. Tee : 36}   & \multicolumn{1}{c|}{M. Shirt : 775}  & \multicolumn{1}{c|}{Light : 180}    \\ \hline
\multicolumn{1}{|c|}{No : 793}             & \multicolumn{1}{c|}{No : 1748}        & \multicolumn{1}{c|}{F. Top : 324}  & \multicolumn{1}{c|}{M. Tee : 793}    & \multicolumn{1}{c|}{Dense : 1748}   \\ \hline
\end{tabular}
\caption{Attribute-wise data description of IIITM Face }
\label{stats-of-database}
\end{table}

% \vspace{-6mm}

Out of these eight attributes described, beard, hair and moustache are subjective and the final label was selected by majority vote across across labels assigned by 5 human annotators. For emotion, the ambiguity was removed by asking the subjects to show a particular emotion and marking it along with the image. The range of emotions were chosen so as to have maximum diversity in terms of how different regions of face are animated for expressing a particular emotion.

\subsection{Problem Formulation} \label{problem-formulation}

Consider $m$ instances of facial images constituting our training samples
where each instance $x_i$ $(i \in \{1, ... , m\})$,  is a colored image of dimension $h \times w$.

$k$ is the total number of attributes and $a_j$  $(j \in \{1,..., k\})$ is the set of valid and mutually exclusive labels for attribute j. The task is to learn a classifier 
\begin{equation}
H: [0, 255]^{h\times w \times 3} \rightarrow \{\{0,1\}^{|a_j|}, j \in \{1,..., k\}\}
\end{equation}

Now, this problem  can be solved in several ways.  In our work, we convert it into a multi-label prediction problem. For doing so, we define $A$ as the union set of all labels occurring in the training samples across all attributes. This is formulated as 
\begin{equation}
A = \bigcup_{j=1}^{k} a_j
\end{equation}

The training labels $y_i$ can be defined as binarized label vectors of length $|A|$ . We further define the label matrix as $\textbf{Y} \in \{0,1\}^{m \times |A|}$ Therefore, the element $y_{ij}$ = 1 or 0 represents whether $jth$ label of $ith$ instance is relevant or not. Finally, the multi-label predictor $H$ becomes 
\begin{equation}
H: [0, 255]^{h\times w \times 3} \rightarrow \{0,1\}^{|A|}
\end{equation}

Another method is to train $k$ different multi-class classification models for each attribute. Hence, $H = \{h_1,...,h_k\}$ and the vector $h_j(x_i)$ will give probability distribution over labels $l \in a_j $.

\subsection{Evaluation Metrics}
Evaluating performance in multi-label classification is more complex as compared to  single-label classification since multiple labels can occur simultaneously in an instance. As a result, we use 9 key metrics that have been proposed in past works focusing on measuring different aspects of multi label classification tasks.

Hammming loss, zero-one loss and coverage loss have been considered in a multitude of works such as \cite{schapire1999boosting, huang2012multi, zhang2014lift}, while F1 score and AUC score in multi label scenario can be extended to instance level (averaging on each instance), micro level (averaging on prediction matrix) and macro level (averaging across each label) metrics. For a unified view on these metrics for multi label classification and their significance, we refer the reader to the work in \cite{wu2017unified}.

\section{Experimental Setup} \label{experimental-setup}

We perform the following experiments on our dataset. Firstly, we train LR, SVM and ResNet-50  referred to as \textbf{LR-MT}, \textbf{SVM-MT} and \textbf{RN-MT} respectively by binarizing the attributes and perform multi-label classification (multi task learning) as described in section \ref{problem-formulation}. Thereafter, we pick the best-performing model in multi-task learning (ResNet in our case, as can be seen in Table \ref{all}), and train 8 such models separately, each for a single attribute. To get a classifier for all attributes, we create an ensemble of these 8 models and refer to it as \textbf{RN-Ens}. Finally, we perform experiments on the IIITM Face-SUE dataset by employing RN-MT classifier which we refer to as \textbf{RN-MT-SUE}. For all the experiments, we keep all images of 85  subjects in the train set and that of 22 subjects in the test set.
%We conduct 2 set of experiments on our new dataset. In the first experiment we train 8 individual classifiers on each of the 8 labels present in our dataset. For this we use a ResNet-50 architecture proposed in \cite{resnet}. This sets a very strong baseline for our multi-label classification problem. In the second experiment we deal with the multi-label classification task as detailed in \ref{problem-formulation}. For this we employ 3 popular classification models - Logistic Regression, Support Vector Machine and ResNet-50. The implementation details are described below

\subsection{Implementation Details}
%%%%%%%%%%%%%%%%%%%%%%%%%%%%%%%%%%%%%%%%%%

%Implementation Details of LR and SVM

%%%%%%%%%%%%%%%%%%%%%%%%%%%%%%%%%%%%%%%%%%
For training the multi-label classification model, we first use Logistic Regression (LR) and Support Vector Machine (SVM) Models. Both of them are trained as One Vs rest Classifiers across the union set of all the labels. The images are resized to $100X100X3$ dimension and then flattened to give a feature vector which is passed to the model. For both LR and SVM models, we have used L2 regularization and C=1, keeping the other hyperparameters with default values as implemented in \cite{scikit-learn}.
The ResNet-50 (hereafter referred to as \textbf{ResNet} only) architecture, wherever used in our experiments, has been constructed as described in \cite{resnet} with some changes as described below. All the images fed to the ResNet classifier are of dimensions $224X224X3$ and the model has been initialized with pretrained weights on ImageNet \cite{deng2009imagenet} following \cite{yosinski2014transferable}. The network is trained using Adam optimization algorithm \cite{kingma2014adam} with a batch size of 64.  Following \cite{lrfind}, we use the LR find mechanism to find initial learning rate and cyclical learning rates for updating the learning rate in subsequent epochs. Note that for all MT experiments, the  background is removed to prevent learning pose from color.

For RN-MT we calculate the hinge loss as we want to predict multiple labels simultaneously using the Equation \ref{hinge-loss}.

\begin{equation} \label{hinge-loss}
    L_i=\sum_{j \neq y_i} max(0,s_{j} - s_{y_i}+ \triangle)
\end{equation}

where $s_{j}$ is the score corresponding to the correct class, $s_{y_i}$ is the score corresponding to the incorrect class and $\triangle$ is the margin. 

For RN-Ens, the loss for each sample $L_i$ is calculated depending on the number of classes $C$ in each label as shown in Equation \ref{losseq}. 

\begin{equation} \label{losseq}
    L_i=
    \begin{cases}
      -(y\log(p) + (1-y)\log(1-p))  & \text{if}\ C=2 \\
      -\sum_{c=1}^{C}y_{o,c}\log{(p_{o,c})}  & C>2
      \end{cases}
\end{equation}

where y (or $y_{o,c}$) indicates binary indicator ($0$ or $1$) if class label $c$ is the correct classification for observation $o$. 

\subsection{Construction of IIITM Face-SUE} \label{IIITM-Face-SUE}
\begin{figure}[H]
\centering
  \begin{subfigure}[b]{0.4\textwidth}
  \includegraphics[width=\textwidth]{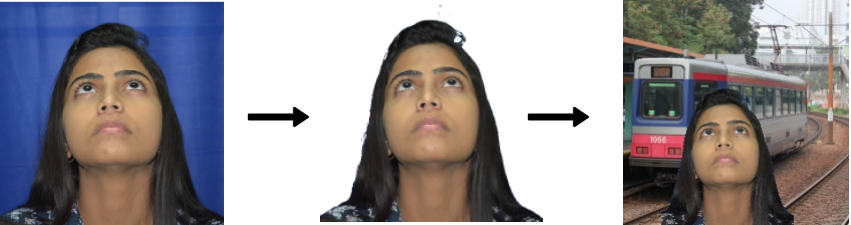}
%   \caption{Steps for simulating unconstrained environment}
    \caption{}
  \label{fig:Ng1} 
\end{subfigure}
\begin{subfigure}[b]{0.4\textwidth}
  \includegraphics[width=\textwidth]{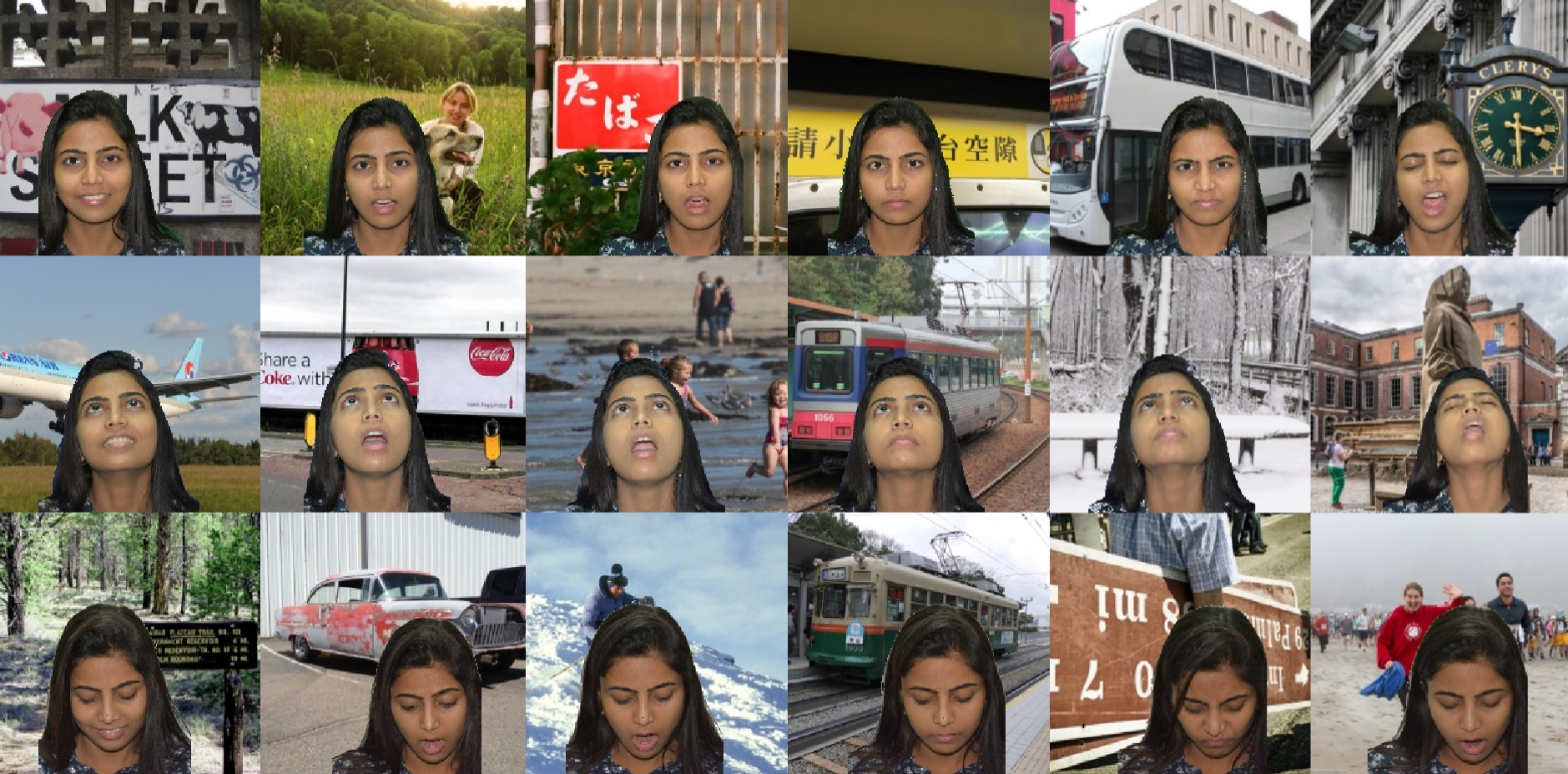}
  \caption{}
  \label{fig:Ng2}
\end{subfigure}
\vspace{-6pt}
\caption{(a) Steps involving the generation of simulated unconstrained environment database. (b) 'In the wild' images for a single subject across all poses and emotions}
\label{fig:aug}
\end{figure}

The original images were captured with a colored background for their easy removal. We use images from \cite{shi2015hierarchical} to replace colored backgrounds with complex scenes (Fig. \ref{fig:aug}). 20\% of these backgrounds were reserved and used for the test set while the remaining were used to distort the images in the train set with the objective that our classifier does not learn to ignore these backgrounds with the same ease as in case of the white background. The subjects were also randomly translated along the X-axis by 5-20\%  to make the task even more challenging. 
% The script to remove the colored background and appending cluttered backgrounds would be released along with the dataset.
\vspace{-1mm}
\vspace{-3mm}

% \begin{figure}[htp]
%     \centering
%     \includegraphics[width=6cm]{process.png}
%     \caption{Attribute diversity in IIITM Face Dataset}
%     \label{fig:galaxy}
% \end{figure}
% \begin{figure}[htp]
%     \centering
%     \includegraphics[width=6cm]{aug_all.png}
%     \caption{Pose and Emotion variation in IIITM Face Dataset }
%     \label{fig:galaxy}
% \end{figure}
The performance of RN-MT-SUE on IIITM Face-SUE has been reported in Table \ref{all}.

\section{Results} \label{sec:results}
\begin{table}[]
\footnotesize
\begin{tabular}{|l|c|c|c|c|c|}
\hline
\textbf{Method} & \multicolumn{4}{c|}{\textbf{Multi-Attribute Detection}} & \textbf{\begin{tabular}[c]{@{}c@{}}Ensemble \\ of Models\end{tabular}} \\ \hline
\textbf{Dataset} & \multicolumn{3}{c|}{\textbf{IIITM Face}} & \textbf{\begin{tabular}[c]{@{}c@{}}IIITM \\ FACE-SUE\end{tabular}} & \textbf{IIITM Face} \\ \hline
\textbf{Metrics} & SVM-MT & LR-MT & RN-MT & RN-MT-SUE & RN-Ens \\ \hline
\textbf{Hamming Loss} & 0.246 & 0.216 & 0.119 & 0.214 & 0.147 \\ \hline
\textbf{Coverage Loss} & 24 & 23.5 & 13.87 & 19.57 & 14.71 \\ \hline
\textbf{Zero-one Loss} & 1 & 0.991 & 0.876 & 0.977 & 0.959 \\ \hline
\textbf{Instance F1} & 0.508 & 0.671 & 0.813 & 0.649 & 0.778 \\ \hline
\textbf{Micro F1} & 0.508 & 0.671 & 0.812 & 0.651 & 0.774 \\ \hline
\textbf{Macro F1} & 0.142 & 0.499 & 0.744 & 0.528 & 0.687 \\ \hline
\textbf{Instance AUC} & 0.661 & 0.753 & 0.887 & 0.760 & 0.863 \\ \hline
\textbf{Micro AUC} & 0.661 & 0.753 & 0.887 & 0.760 & 0.863 \\ \hline
\textbf{Macro AUC} & 0.5 & 0.636 & 0.839 & 0.697 & 0.811 \\ \hline
\end{tabular}
\caption{Benchmarking performance in face attribute classification on IIITM Face using different models}
\label{all}
\vspace{-8mm}
\end{table}

\begin{figure}[htp]
    \centering
    \includegraphics[width=8cm]{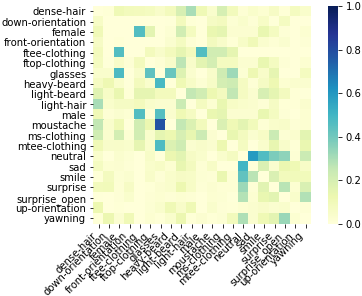}
    \vspace{-4pt}
    \caption{The upper(lower) triangular matrix is the absolute difference between correlation of true labels and the predicted labels of RN-MT(RN-Ens). Dark blue cells indicate  the failure at capturing the underlying correlation between attributes. }
    \label{fig:heatmap}
\end{figure}

We make several key observations based on the results shown in Table \ref{all}. 
% Comment on performance of the ResNet model as compared to LR and SVM.
% The performance of individual classifiers of RN-Ens on IIITM Face shown in Table \ref{} can serve as a benchmark for researchers attempting to deal with individual attributes like emotion, orientation etc.
Firstly, our experiments show that ResNet is far more superior to LR and SVM when it comes to multi-attribute classification on image datasets.
Moreover, it can be seen that RN-MT outperforms RN-Ens across all metrics of evaluation. We attribute this to the ability of RN-MT to learn correlation among various attributes whereas the ensemble model is not presented with this opportunity since all the individual classifiers are trained in isolation. Our claims are bolstered by the fact that the zero-one loss is substantially high in case of RN-Ens. Zero-one loss being a measure of absolute correctness across all the attributes signifies that the our single model is able to capture this correlation surprisingly well. 
The heatmap in Fig. \ref{fig:heatmap} represents how close is the correlation between the true labels as compared to the correlation between predicted labels for both RN-Ens and RN-MT. A smaller value (light yellow cell) indicates that the model has learnt the underlined relationship between the labels. 
Further, the RN-Ens model, being an ensemble, was seen to be computationally expensive at evaluation time.
% First we calculate the correlation between all pairs of labels for the predicted labels of both RN-MT and RN-Ens as well as the correlation across all pairs of true labels.
\vspace{-3mm}
\vspace{-10mm}
Our experiment to simulate `in the Wild' behavior using IIITM Face-SUE shows that there is a drastic degradation in performance of RN-MT-SUE classifier across all metrics. We attribute this to ResNet's ability to easily filter out white backgrounds as compared to the complex scenes.

\section{Future Works}

The release of IIITM Face presents several opportunities to the research community for future exploration. Being a high resolution image dataset, it can be used to evaluate the performance of models trained on low resolution images, on a high resolution dataset such as this one and vice-versa. Moreover, it can serve as a test-bed for evaluation of state-of-the-art techniques in emotion recognition, specifically in the Indian context. Additionally, the subsequent conversion to an `in the Wild' dataset  opens the doors for evaluation of state-of-the-art techniques in face-attribute detection as well. In future works, we also intend to study activation maps of model to reveal where it is looking to get information about specific attributes.
%Also as stated in Section \ref{sec:results} one can observe the activation maps of Conv-Nets to study the impact of noise and cluttered background task. We aim to study several of these points in our future works. % Improve this line.
\pagebreak

\begin{acks}
This work was supported by the Department of Science and
Technology, New Delhi, India under Technology System
Development Program project - DST/TSG/NTS/2013/19-G. We would like to thank the Library Staff, PH.D scholars and students at IIITM Gwalior for patiently posing in front of the camera.
\end{acks}

\bibliographystyle{ACM-Reference-Format}
\bibliography{acmart.bib}


\end{document}